# Using T-norm Based Uncertainty Calculi in a Naval Situation Assessment Application

Piero P. Bonissone

General Electric Corporate Research and Development
1 River Road, K1-5C32A, Schenectady, New York 12301
Phone: 518-387-5155 Arpanet: Bonissone@GE-CRD

## ABSTRACT

RUM (Reasoning with Uncertainty Module), is an integrated software tool based on a KEE, a frame system implemented in an object oriented language. RUM's architecture is composed of three layers: *representation, inference*, and *control.*

The representation layer is based on frame-like data structures that capture the uncertainty information used in the inference layer and the uncertainty meta-information used in the control layer. The inference layer provides a selection of five T-norm based uncertainty calculi with which to perform the intersection, detachment, union, and pooling of information. The control layer uses the meta-information to select the appropriate calculus for each context and to resolve eventual ignorance or conflict in the information. This layer also provides a context mechanism that allows the system to focus on the relevant portion of the knowledge base, and an uncertain-belief revision system that incrementally updates the certainty values of well-formed formulae (*wffs*) in an acyclic directed deduction graph.

RUM has been tested and validated in a sequence of experiments in both naval and aerial situation assessment (SA), consisting of correlating reports and tracks, locating and classifying platforms, and identifying intents and threats. An example of naval situation assessment is illustrated. The testbed environment for developing these experiments has been provided by LOTTA, a symbolic simulator implemented in Zetalisp Flavors. This simulator maintains time-varying situations in a multi-player antagonistic game where players must make decisions in light of uncertain and incomplete data. RUM has been used to assist one of the LOTTA players to perform the SA task.

### 1. INTRODUCTION

The trend followed by most approaches for reasoning with uncertainty has shown an almost complete disregard for the fundamental issues of automated reasoning, such as the proper *representation* of information and meta-information, the allowable *inference* paradigms suitable for the representation, and the efficient *control* of such inferences in an explicitly programmable form. The majority of the approaches to reasoning with uncertainty do not properly cover these issues. Some approaches lack expressiveness in their representation paradigm. Other approaches require unrealistic assumptions to provide uniform combining rules defining the plausible inferences. Most approaches do not even recognize the need for having an explicit control of the inferences.

This lack of awareness has been the driving force for compiling a list of requirements (desiderata) that each reasoning system handling uncertain information should satisfy. Following the typical structure of automated reasoning techniques, the list of requirements has been organized in three layers: *representation, inference*, and *control*. The extension of this explicit layered separation from *crisp*-reasoning systems to *uncertain*-reasoning systems is a natural step leading to a better integration of the management of uncertainty with the various techniques for automated reasoning.

* This work was partially supported by the Defense Advanced Research Projects Agency (DARPA) under USAF/Rome Air Development Center contract F30602-85-C-0033. Views and conclusions contained in this paper are those of the authors and should not be interpreted as representing the official opinion or policy of DARPA or the U.S. Government.

An in-depth treatment of the layered desiderata can be found in a previous paper [5]. In this article we describe RUM (Reasoning with Uncertainty Module), which represents our answer to the desiderata. We also illustrate a naval situation assessment problem which is used to validate RUM. This application is based on an architecture designed to simulate various military scenarios involving Multi-Sensors/Multi-Targets (MS/MT) and to perform situation assessment (SA) related tasks. The MS/MT architecture, illustrated in Figure 1, is composed of two major blocks: a reasoning system and a simulation environment.

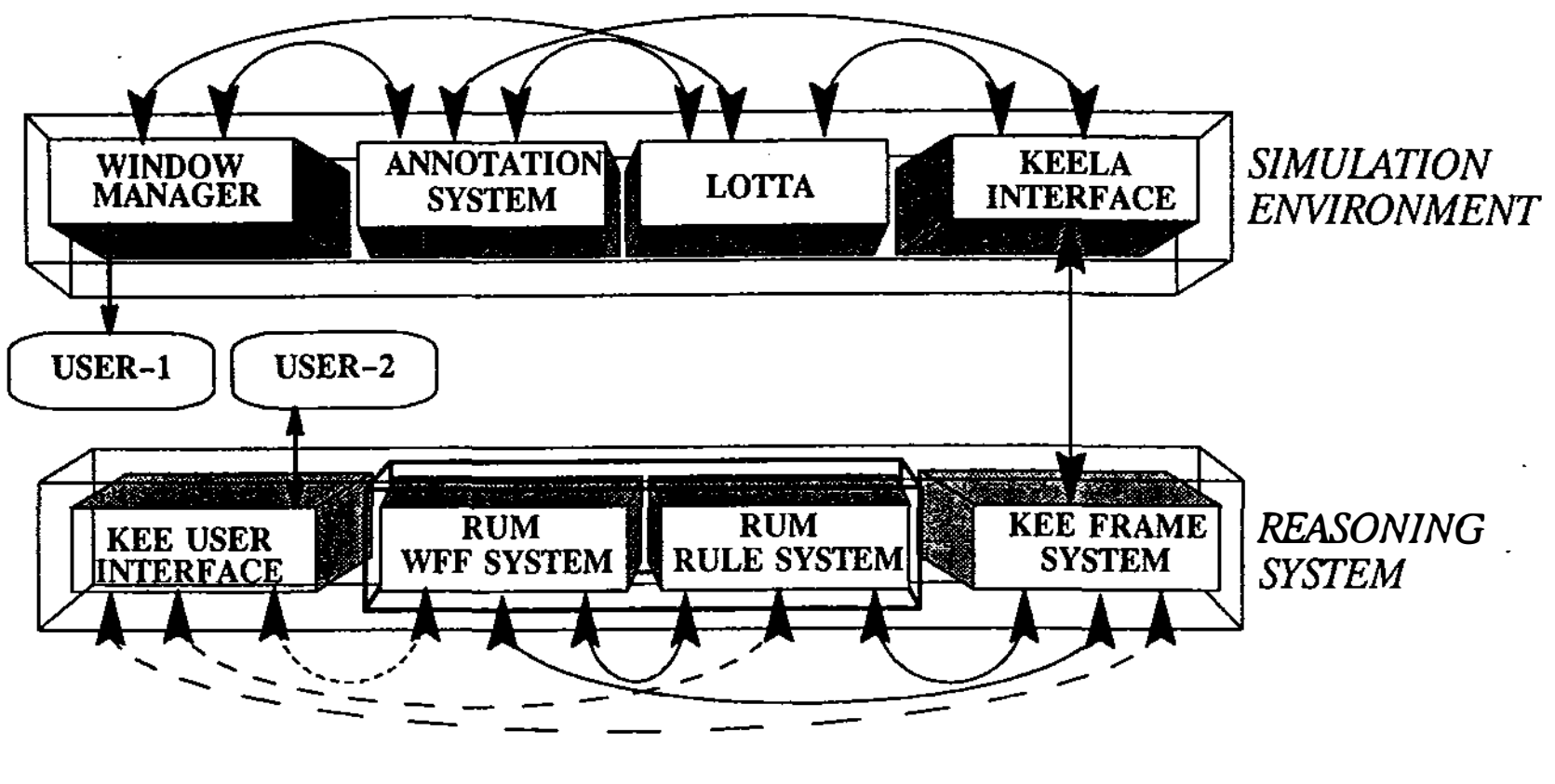

**Figure 1: Architecture for Multi-Sensors/Multi-Targets (MS/MT)**

RUM is the reasoning system used in this architecture. This system, built according to the three layer desiderata, is thoroughly described in [3]. It is summarized in section 2, with a particular focus on its control layer. The second block of the MS/MT architecture, the simulation environment, is described in section 3, in conjunction with some definitions of the tasks required to perform situation assessment. The last two sections contain an analysis of the MS/MT experiment and some preliminary conclusions on this work.

## 2. RUM, THE REASONING SYSTEM

RUM is an integrated software tool based on KEE[T.M.], a frame system implemented in an object oriented language. The underlying theory of RUM, centered around the concept of Triangular norms, was described in two previous articles [2,5]. RUM's architecture is composed of three layers: *representation, inference,* and *control.* A philosophical motivation for RUM's three layer organization can be found in [3]. This section summarizes some of the theoretical results and provides a unified framework for their interpretation and use in RUM's architecture.

### 2.1 Representation: the Wff System and the Rule Language

The representation layer is based on frame-like data structures that capture the uncertainty information used in the inference layer and the uncertainty meta-information used in the control layer.

#### 2.1.1 RUM's Wff System

RUM's Wff System modifies KEE's representation of a *wff* (well-formed formula). RUM's *wff* is the pair [*<unit>* *<slot>*], which is the description of a variable in the problem domain. For each *wff* a corresponding uncertainty unit is created. The unit contains a list of the values that were considered for the *wff.* For each value the unit maintains its certainty's *lower* and *upper bounds*, an *ignorance measure*, a *consistency measure*, and the *evidence source*.

Figure 2 illustrates an example of an uncertainty unit attached to a *wff.* The *wff* is the variable [*Platform-439 Classs-name*]. In the uncertainty unit, under the slot VALUES, we can see the possible values which were considered by the system and their corresponding certainty bounds. The uncertainty unit also maintains a record of the rule instances which were fired to derive such values (for inferred *wffs*, this logical support represents the evidence source).

RUM's Wff System allows the user to express arbitrary uncertainty granularity by providing the flexibility to mix precise and imprecise measures of certainty in defining the input certainty (points, intervals, fuzzy numbers/intervals, linguistic values) and the rule strengths (categorical and plausible

IF/IFF). Various term sets of linguistic probabilities with fuzzy-valued semantics [2] provide a selection of input granularity. The values of the terms can be used as default values or can be modified by the user.

```
Own slot: CONSTRAINTS.RULES from PLATFORM-439-CLASS.NAME
   Inheritance: OVERRIDE.VALUES
   Values: FISHING.BOAT.POS.ID-OK-300-TRACK-3 in MSMT,
           FISHING.BOAT.POS.ID-OK-300-TRACK-0 in MSMT

Own slot: DEPENDENT.RULES from PLATFORM-439-CLASS.NAME
   Inheritance: OVERRIDE.VALUES
   ValueClass: GENERIC.RULE.UNIT in HARDWARE
   Comment: another list of rules.
   Values: MERCHANT.TYPE.LINK-1000-PLATFORM-439 in MSMT,
           FISHING.TYPE.LINK-1010-PLATFORM-439 in MSMT

Own slot: DS.RULES from PLATFORM-439-CLASS.NAME
   Inheritance: OVERRIDE.VALUES
   ValueClass: GENERIC.RULE.UNIT in HARDWARE
   Avunits: AV.BAD in HARDWARE
   Comment: Rules to be combined using the Dempster-Schaefer sonorm.
   Values: UNKNOWN

Own slot: FLAG from PLATFORM-439-CLASS.NAME
   Inheritance: OVERRIDE.VALUES
   Avunits: AV.FLAG in HARDWARE, AV.ALERT in HARDWARE
   Cardinality.Min: 1
   Cardinality.Max: 1
   Comment: Good or X.
   Values: GOOD

Own slot: NECESSITY from PLATFORM-439-CLASS.NAME
   Inheritance: OVERRIDE.VALUES
   Avunits: AV.POLLUTE in HARDWARE
   Cardinality.Min: 1
   Cardinality.Max: 1
   Comment: Minimum support for a wff.
   Values: (0.6928104 0.816568 0.08169937 0.07843958)

Own slot: PLAUSIBILITY from PLATFORM-439-CLASS.NAME
   Inheritance: OVERRIDE.VALUES
   Avunits: AV.POLLUTE in HARDWARE
   Cardinality.Min: 1
   Cardinality.Max: 1
   Comment: Maximum support for a wff.
   Values: 1

Own slot: S1.RULES from PLATFORM-439-CLASS.NAME
   Inheritance: OVERRIDE.VALUES
   ValueClass: GENERIC.RULE.UNIT in HARDWARE
   Avunits: AV.BAD in HARDWARE
   Comment: Rules that are to be disjuncted using S1.
   Values: UNKNOWN

Own slot: S15.RULES from PLATFORM-439-CLASS.NAME
   Inheritance: OVERRIDE.VALUES

Own slot: S2.RULES from PLATFORM-439-CLASS.NAME
   Inheritance: OVERRIDE.VALUES
   ValueClass: GENERIC.RULE.UNIT in HARDWARE
   Avunits: AV.BAD in HARDWARE
   Comment: Rules to be disjuncted using S2.
   Values: SUB.POS.ID-CLOSE.POP.UP-500-TRACK-3 in MSMT,
           MERCHANT.NEG.ID-MANEUVERS-210-PLATFORM-439 in MSMT,
           MERCHANT.NEG.ID-DISTANT.POP.UP-220-TRACK-3 in MSMT,
           MERCHANT.NEG.ID-DISTANT.POP.UP-220-TRACK-0 in MSMT,
           SUB.POS.ID-CLOSE.POP.UP-500-TRACK-0 in MSMT,
           SUB.POS.ID-SONAR-650-TRACK-3 in MSMT,
           SUB.POS.ID-SONAR-650-TRACK-0 in MSMT

Own slot: S25.RULES from PLATFORM-439-CLASS.NAME
   Inheritance: OVERRIDE.VALUES
   ValueClass: GENERIC.RULE.UNIT in HARDWARE
   Avunits: AV.BAD in HARDWARE
   Comment: Rules to be disjuncted using S2.5.
   Values: FISHING.BOAT.NEG.ID-USE.RADAR-330-TRACK-0 in MSMT,
           FISHING.BOAT.NEG.ID-TOO.FAST-310-TRACK-3 in MSMT,
           MERCHANT.POS.ID-OK-100-TRACK-0 in MSMT,
           MERCHANT.POS.ID-OK-100-TRACK-3 in MSMT,
           MERCHANT.NEG.ID-TOO.SLOW-140-TRACK-3 in MSMT,
           MERCHANT.NEG.ID-TOO.SLOW-140-TRACK-0 in MSMT,
           MERCH.NEG.ID-TOO.SMALL-200-TRACK-3 in MSMT,
           MERCH.NEG.ID-TOO.SMALL-200-TRACK-0 in MSMT,
           FISHING.BOAT.NEG.ID-TOO.FAST-310-TRACK-0 in MSMT,
           FISHING.BOAT.NEG.ID-USE.RADAR-330-TRACK-3 in MSMT

Own slot: S3.RULES from PLATFORM-439-CLASS.NAME
   Inheritance: OVERRIDE.VALUES
   ValueClass: GENERIC.RULE.UNIT in HARDWARE
   Avunits: AV.BAD in HARDWARE
   Comment: Rules to be disjuncted using S3.
   Values: FISHING.BOAT.NEG.ID-TOO.FAR-340-TRACK-0 in MSMT,
           FISHING.BOAT.NEG.ID-TOO.FAR-340-TRACK-3 in MSMT,
           FISHING.BOAT.NEG.ID-TOO.BIG-320-TRACK-3 in MSMT,
           MERCHANT.NEG.ID-DODGE.STATIC.SENSOR-250-PLATFORM-439 in MSMT,
           MERCHANT.NEG.ID-BAD.WEATHER-260-PLATFORM-439 in MSMT,
           FISHING.BOAT.NEG.ID-TOO.BIG-320-TRACK-0 in MSMT

Own slot: VALUE from PLATFORM-439-CLASS.NAME
   Inheritance: OVERRIDE.VALUES
   Comment: Value of slot.
   Values: (SUBMARINE MERCHANT FISHING.BOAT),
           (((0 0 0 0) (0.19999999 0.37 0.06 0.05))
                  ((0.6928104 0.816568 0.08169937 0.07843958) (1 1 0 0))
                  ((0 0 0 0) (0.019801915 0.039215684 0.019802034 0.09162535)))
```

**Figure 2: Uncertainty Unit Associated with wff** [*Platform-439 Class-name*]

### 2.1.2 RUM's Rule System: The Rule Language

RUM's Rule System replaces KEE Rule System-3 capabilities by incorporating uncertainty information in the inference scheme. The uncertain information is described in the uncertainty units of the *wffs*, represented in RUM's Wff System, and in the degrees of sufficiency and necessity attached to each rule*. The degree of sufficiency denotes the extent to which one should believe in the rule conclusion, if the rule premise is satisfied. The degree of necessity indicates the confidence with which one can negate the conclusion, if the premise fails.

A rule is internally represented by a frame with several slots. These slots include the name of the rule; the lists of contexts, premises, and conclusions; the rule's sufficiency and necessity; and the T-norm to be used for aggregation. All slots (except the name, premises, and consequences) have default values. The contexts, premises, and conclusions can comprise values, variables, RUM predicates and arbitrary LISP functions. Rules with unbound variables are instantiated with the necessary environment to produce rule instances. An example of two RUM rules is provided in section 3.2.

The T-norm specified with each rule is used to aggregate the certainties of the rule premises and to perform detachment (which computes the certainty of the conclusion given the sufficiency and necessity of the rule). It defaults to $T_3$, which is the MIN function. The associated T-conorm is used to aggregate the certainties of identical conclusions inferred by multiple rule instances derived from the *same* rule. These are often subsumptive, and the value defaults to $S_3$, the MAX function. Finally, each separate consequence of a rule has a specified T-conorm that will be used to aggregate the consequence with identical consequences derived from *different* rules. (i.e., multiple assignments of the same value to the wff). The negation operator causes the wff to be assigned the complemented value.**

* It is important to note that the inference symbol $\rightarrow$ in the production rule $A \rightarrow^s B$ is interpreted as a (weak) ***material implication*** operator in multiple-valued logics. The value $s$ is the lower bound of the degree of sufficiency of the implication. This is in contrast with the interpretation of ***conditioning***, i.e., $s$ = P(B|A). The symbol $\leftrightarrow$ in the production rule $A \leftrightarrow^{s,n} B$ is interpreted as a (weak) ***logical equivalence*** operator in multiple-valued logics, in which $s$ and $n$ are the lower bounds of sufficiency and necessity, respectively. This (weak) logical equivalence is an *if-and-only-if* (IFF) rule, which can be decomposed into the two rules: $A \rightarrow^s B$ and $B \rightarrow^n A$ (equivalent to $\neg A \rightarrow^n \neg B$). RUM's rules are of the type $C \rightarrow (A \leftrightarrow^{s,n} B)$, where C indicates the context of the rule (see section 2.3.3) and $\rightarrow$ represents the strong material implication.

** If a *wff* has a value A with an If the certainty interval attached to a value A is [L(A), U(A)], its complemented value, ¬A, has a certainty interval defined by [1-U(A), 1-L(A)].

### 2.2 Inference: Triangular norms (T-norms) Based Calculi

The inference layer is built on a set of five Triangular norms (T-norms) based calculi. The T-norms' associativity and truth functionality entail problem decomposition and relatively inexpensive belief revision. The theory of T-norms has been covered in previous articles [2,5]. A brief review of their definition and their use in RUM is included for the reader's convenience.

#### 2.2.1 Background Information on T-norms

Triangular norms (T-norms) and Triangular conorms (T-conorms) are the most general families of binary functions that satisfy the requirements of the conjunction and disjunction operators, respectively. T-norms and T-conorms are two-place functions from [0,1]x[0,1] to [0,1] that are monotonic, commutative and associative. Their corresponding boundary conditions, i.e., the evaluation of the T-norms and T-conorms at the extremes of the [0,1] interval, satisfy the truth tables of the logical AND and OR operators.

In a previous paper [2], six parametrized families of T-norms and dual T-conorms were discussed and analyzed by the author. Of the six parametrized families, one family was selected due to its complete coverage of the T-norm space and its numerical stability. This family, originally defined by Schweizer & Sklar [10], was denoted by $T_{Sc}(a,b,p)$, where $p$ is the parameter that spans the space of T-norms. More specifically:

$$T_{Sc}(a,b,p) = (a^{-p} + b^{-p} - 1)^{-\frac{1}{p}} \quad \text{if } (a^{-p} + b^{-p}) \geq 1 \quad \textit{when } p < 0$$

$$T_{Sc}(a,b,p) = 0 \quad \text{if } (a^{-p} + b^{-p}) < 1 \quad \textit{when } p < 0$$

$$T_{Sc}(a,b,0) = \lim_{p \to 0} T_{Sc}(a,b,p) = ab \quad \textit{when } p \to 0$$

$$T_{Sc}(a,b,p) = (a^{-p} + b^{-p} - 1)^{-\frac{1}{p}} \quad \textit{when } p > 0$$

Its corresponding T-conorm, denoted by $S_{Sc}(a,b,p)$, was defined as:

$$S_{Sc}(a,b,p) = 1 - T_{Sc}(1-a,\, 1-b,\, p)$$

In the same paper it was shown that the use of term sets determines the granularity with which the input certainty is described. This granularity limits the ability to differentiate between two similar calculi; the numerical results obtained by using two calculi whose underlying T-norms are very close in the T-norm space will fall within the same granule in a given term set. Therefore, only a finite, small subset of the infinite number of calculi that can be generated from the parametrized T-norm family produces notably different results. The number of calculi to be considered is a function of the uncertainty granularity.

This result was confirmed by an experiment [2] where eleven different calculi of uncertainty, represented by their corresponding T-norms, were analyzed. To generate the eleven T-norms, the parameter $p$ in Schweizer's family was given the following values:

-1, -0.8, -0.5, -0.3, 0 (in the limit), 0.5, 1, 2, 5, 8, and $\infty$ (in the limit).

The experiment showed that five equivalence classes were needed to represent (or reasonably approximate) any T-norm, when term sets with at most thirteen elements were used. The corresponding five uncertainty calculi were defined by the common negation operator $N(a) = 1-a$ and the DeMorgan pair $(T_{Sc}(a,b,p), S_{Sc}(a,b,p))$ for the following values of p:

| | | |
|---|---|---|
| $p = -1$ | $T_1(a,b) = max(0,\, a+b-1)$ | $S_1(a,b) = min(1,\, a+b)$ |
| $p = -0.5$ | $T_{Sc}(a,b,-0.5) = \max(0,\, a^{0.5}+b^{0.5}-1)^2$ | $S_{Sc}(a,b,-0.5) = 1-\max(0,[(1-a)^{0.5}+(1-b)^{0.5}-1])^2$ |
| $p \to 0$ | $T_2(a,b) = ab$ | $S_2(a,b) = a + b - ab$ |
| $p = 1$ | $T_{Sc}(a,b,1) = \max(0,\, a^{-1}+b^{-1}-1)^{-1}$ | $S_{Sc}(a,b,1) = 1-\max(0,[(1-a)^{-1}+(1-b)^{-1}-1])^{-1}$ |
| $p \to \infty$ | $T_3(a,b) = min(a,\, b)$ | $S_3(a,b) = max(a,\, b)$ |

RUM's inference layer provides the user with a selection of the five T-norm based calculi described above. They are referred to as $T_1$, $T_{1.5}$, $T_2$, $T_{2.5}$, $T_3$, respectively.

**2.2.2 Operations in a T-norm Based Calculus**

For each calculus, four operations are defined in RUM's Rule System: *premise evaluation, conclusion detachment, conclusion aggregation,* and *source consensus.* Each operation in a calculus can be completely defined by a Triangular norm *T(.,.)*, and a negation operator *N(.)*, just as in classical logic any boolean expression can be rewritten in terms of an intersection and complementation operator. A formal justifications for the following definitions can be found in Reference 4. The four operations are defined as follows:

*Premise evaluation:* The premise evaluation operation determines the degree to which all the clauses in the rule premise have been satisfied by the matching *wffs*. Let $b_i$ and $B_i$ indicate the lower and upper bounds of the certainty of condition $i$ in the premise of a given rule. Then the premise certainty range [b,B] is defined as:

$$[b,B] = [\, T(b_1, b_2, \ldots, b_m), T(B_1, B_2, \ldots, B_m) \,]$$

*Conclusion Detachment:* The conclusion detachment operation indicates the certainty with which the conclusion can be asserted, given the strength and appropriateness of the rule. Let $s$ and $n$ be the lower bounds of the degree of *sufficiency* and *necessity*, respectively, of the given rule, and let [b,B] be the computed premise certainty range. Then the range [c,C], indicating the lower and upper bound for the certainty of the conclusion inferred by such rule, is defined as:

$$[c,C] = [\, T(s, b), N(\, T(n, N(B))) \,]$$

The degrees of sufficiency and necessity respectively indicate the amount of certainty with which the rule premise implies its conclusion and viceversa. The sufficiency degree is used with *modus ponens* to provide a lower bound of the conclusion. The necessity degree is used with *modus tollens* to obtain a lower bound for the complement of the conclusion (which can be transformed into an upper bound for the conclusion itself).

*Conclusion aggregation:* The conclusion aggregation operation determines the consolidated degree to which the conclusion is believed if supported by more than one path in the rule deduction graph, i.e. by more than one rule instance. It is also possible to have various groups of deductive paths, i.e. various sets of rule instances, all supporting the same conclusion. Each group of deductive paths can have a distinct conclusion aggregation operator associated with it. Let the ranges $[c_j, C_j]$ indicate the certainty lower and upper bounds of the *same* conclusion inferred by various rules instances belonging to the same group. Then, for each group of deductive paths, the range [d,D] of the aggregated conclusion is defined as:

$$[d, D] = [\, N(\, T(N(c_1), N(c_2), \ldots, N(c_m)), T(N(C_1), N(C_2), \ldots, N(C_m))) \,]$$

RUM distinguishes between rule instances generated from the same rule and rule instances derived from different rules. The first type or rule instances is aggregated first, to take into account the usually large amount of redundancy that such rule instances entail. The second set of rule instances is subsequently aggregated taking into account the knowledge about the presence or lack of positive/negative correlation that characterizes the various rules.

*Source Consensus:* The source consensus operation reflects the fusion of the certainty measures of the same evidence A provided by different sources. The evidence can be an *observed* fact, or a *deduced* fact. In the former case, the fusion occurs before the evidence is used as an input in the deduction process. In the latter case, the fusion occurs after the evidence has been aggregated by each group of deductive paths. The source consensus operation reduces the ignorance about the certainty of A, by producing an interval that is always smaller or equal to the smallest interval provided by any of the information source. If there is an inconsistency among some of the sources, the resulting certainty intervals will be disjoint, thus introducing a conflict in the aggregated result. Let $[L_1(A), U_1(A)]$, $[L_2(A), U_2(A)]$, ..., $[L_n(A), U_n(A)]$ be the certainty lower and upper bounds of the same conclusion provided by different sources of information. Then, the result $[L_{tot}(A), U_{tot}(A)]$, obtained from *fusing* all the assertions about A, is given by taking the intersection of the certainty intervals:

$$[L_{tot}(A), U_{tot}(A)] = [\, Max_i\ L_i(A), Min_i\ U_i(A)]$$

### 2.3 Control: Calculus selection, Uncertain-Belief Revision, Context Mechanism

#### 2.3.1 Calculi Selection

As it was discussed in the previous section, RUM's Rule System uses a set of five T-norm based calculi. The calculus used by each rule instance is inherited from its rule subclass (the rule before the instantiation). The calculus can be modified through KEE's user interface or programmatically (i.e. by an active value). Class inheritance can also be used to modify the degree of sufficiency and necessity of all the rule members of the same class.

The calculi selection consists of two assignments. The first assignment indicates the T-norm with which the premise evaluation and the conclusion detachment will be computed. Such an assignment is made for each rule, and, through inheritance, is passed to all rule instances derived from the same rule.

The second assignment indicates the T-conorm (represented by its dual T-norm) with which the conclusion aggregation will be computed. This assignment is made for each subset of rule instances generated from *different* rules and asserting the same conclusion.

##### 2.3.1.1 Rationale for Calculi Selection

The T-norm characteristics will determine the selection choices. For the first assignment, the T-norm assigned to each rule for the premise evaluation and the conclusion detachment will be a function of the decision maker's *attitude toward risk*. The ordering of the T-norms, which is identical to the ordering of parameter $p$ in the Schweizer & Sklar family of T-norms, reflects the ordering from a conservative attitude ($p=-1$ or $T_1$) to a non-conservative one ($p \rightarrow \infty$ or $T_3$). From the definition of the calculi operations, we can see that $T_1$ will generate the smallest premise evaluation and the weakest conclusion detachment (i.e., the widest uncertainty interval attached to the rule's conclusion). T-norms generated by larger values of $p$ will exhibit less drastic behaviors and will produce nested intervals with their detachment operations. $T_3$ will generate the largest premise evaluation and the strongest conclusion detachment (the smallest certainty interval).

For the second assignment, the T-norm assigned to the subsets of rule instances (derived from different rules and asserting the same conclusion) will be a function of the *lack or presence of positive/negative correlation* among the rules in each subset. The ordering of the T-norms reflects the transition from the case of extreme negative correlation, i.e., mutual exclusiveness ($T_1$), through the case of uncorrelation ($T_2$), to the case of extreme positive correlation, i.e., subsumption ($T_3$).

Currently, all calculi assignments are explicitly made and modified through the user interface, to exercise the implemented accessing functions. In the next development phase of RUM control layer, the calculi assignments will be made by a set of selection rules expressing the meta-knowledge about the context. These rules will select the T-norms that better reflect the knowledge engineer's desired attitude toward risk and the perceived amount of correlation among the rules used in such a context.

#### 2.3.2 Uncertain-Belief Revision

A daemon-based implementation of the belief revision of the uncertain information is available in the control layer of RUM's Rule System. For any conclusion made by a rule, the belief revision mechanism monitors the changes in the certainty measures of the *wffs* that constitute the conclusion's support or the changes in the calculus used to compute the conclusion certainty measure. Validity flags are inexpensively propagated through the rule deduction graph. Five types of flag values are used:

**Good** Guarantees the validity of the cached certainty measure detached by the rule instance and aggregated into the associated wff.

**Bad (level i)** Indicates that the cached certainty measure detached by the rule instance is no longer reliable, since the support of some of the wff's in the premise of this rule instance has changed. The *i*th level indicates the correct order of recomputation.

**Inconsistent** Indicates that the cached certainty measure associated with the wff is conflicting. The inconsistency can be removed by executing a locally defined procedure (differential diagnosis type of experiment, recency of information, split in possible words with subsets of the original sources, etc.)

**Not Applicable** Indicates that the context of the rule instance is no longer active and the rule instance contribution to the aggregated certainty measure of the wff should be ignored.

**Ignorant** Indicates that the cached certainty measure detached by the rule instance is too vague to be useful. The default behavior is to ignore the rule instance contribution to the aggregated certainty measure of the wff. Locally defined procedure could be used to remove the ignorance if so specified.

#### 2.3.2.1 An Example of Using the Uncertain-Belief Revision

To provide the reader with a better understanding of the uncertain-belief revision, we will make the following graphical analogy: the *wffs* of the reasoning system correspond to nodes in an acyclic deductive graph; the inference rules in the system correspond to the inference gates that connect the nodes in the graph. There are two types of *wffs*: the observations or assumptions, corresponding to the nodes at the ***frontier*** of the graph, and the inferred conclusions, corresponding to the ***intermediate*** nodes in the graph. The first type of node does not have any logical support (its evidence source is the observer or the assumption's maker). The second type of node has a logical support represented by the set of rule instances that made that inference. For this second type, the logical support is the evidence source.

Figure 3 illustrates a a portion of an acyclic deductive graph, in which rule instances are depicted as gates. In the graph we can observe the following five rules:

| | | | | | | |
|---|---|---|---|---|---|---|
| R1: | C → | (A,B ↔ J) | suffic. = $s_1$ | necess. = $n_1$ | calcul. = $T_2$ | aggreg. = $S_2$ |
| R2: | C → | (D ↔ J) | suffic. = $s_2$ | necess. = $n_2$ | calcul. = $T_2$ | aggreg. = $S_2$ |
| R3 | | (E ↔ J) | suffic. = $s_3$ | necess. = $n_3$ | calcul. = $T_3$ | aggreg. = $S_3$ |
| R4: | H → | (E,F,G ↔ J) | suffic. = $s_4$ | necess. = $n_4$ | calcul. = $T_3$ | aggreg. = $S_3$ |
| R5 | | (J,I ↔ K) | suffic. = $s_5$ | necess. = $n_5$ | calcul. = $T_3$, | aggreg. = $S_2$ |

Two more rules, R6 and R7, are partially shown in the same figure.

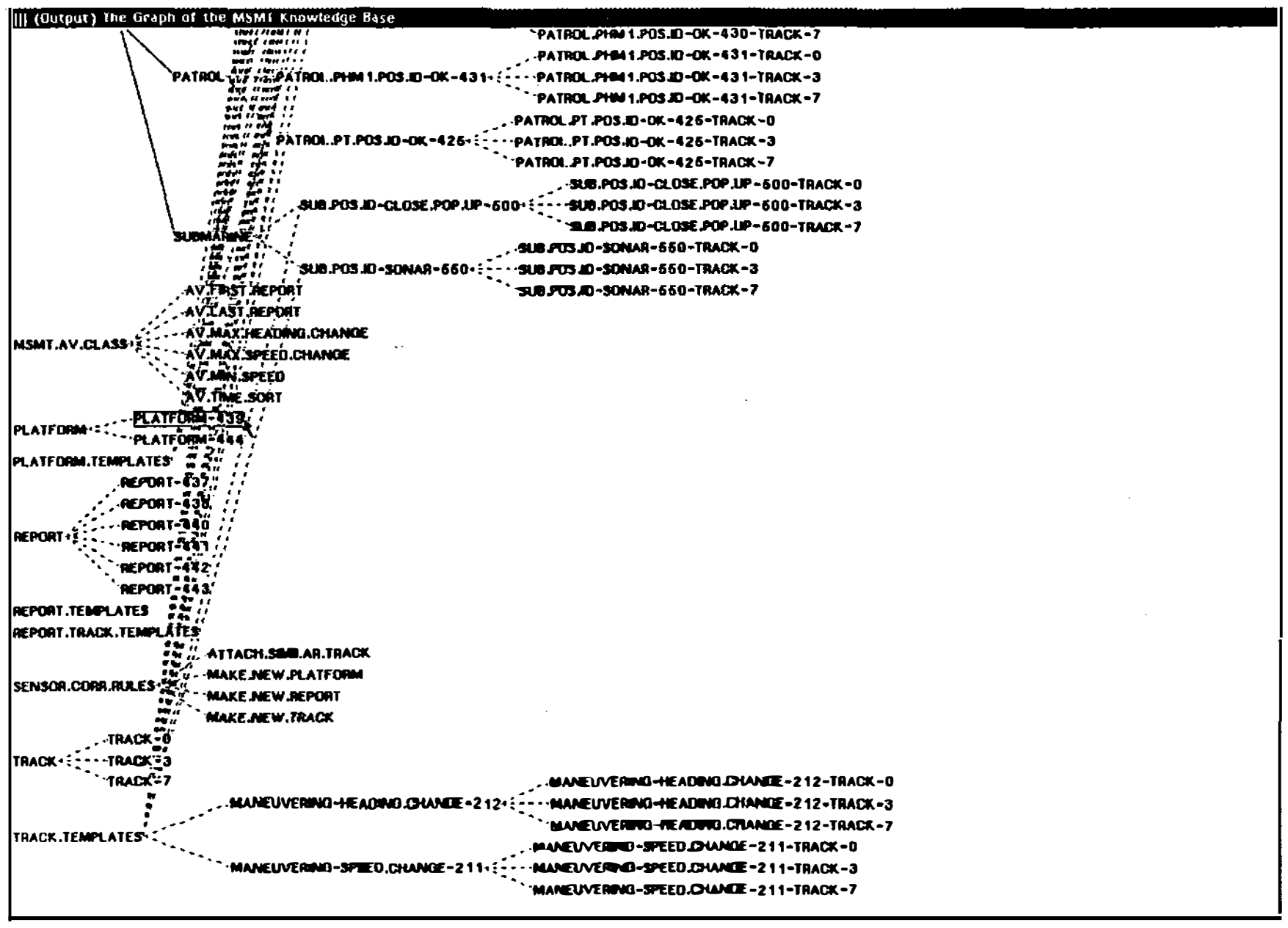

**Figure 3: Portion of an Acyclic Deductive Graph**

In Figure 3, C and H (depicted as control lines on the side of a gate) represent two context descriptions that enable/disable the activation of rules R1, R2, R4. The other two rules (R3 and R5) are always potentially active (regardless of context). The figure shows the case in which fact D has just changed. This change causes the propagation of a bad-validity flag that affects the conclusion of rules R2 and R5 (J and K, respectively). The numbers attached to the bad flag indicate the order in which a recomputation of the certainty measures must be performed. Fact H has also changed and its new value no longer satisfies the context description of rule R4, thus causing the not-applicable flag to be attached to the detachment of R4. Fact L has also changed, affecting the validity of Rule R6's detachment.

#### 2.3.2.2 Reasoning under Pressure

The belief revision system offers both backward and forward processing. Running in *depth-first, backward mode*, RUM recomputes the certainty measures of the modified *wffs* that are required to answer a given query. This mode (called reasoning under pressure) is used when the system or the user decide that they are dealing with time-critical tasks. In the case illustrated in the previous figure, if the value of *wff* K were requested, the systems would perform the following sequence of tasks: fetch the new certainty values of D (lower and upper bounds); recompute the detachment of rule R2; use T-conorm $S_2$ to evaluate the OR node (with R1 and R2's detachments); ignore R4's detachment, treating R3's detachment as the only input to the OR node associated with T-conorm $S_3$; fuse the two OR nodes, defining the new certainty values of *wff* J; recompute the detachment of rule R5; use T-conorm $S_2$ to evaluate the OR node (with R5 and R7's detachments), defining the new certainty values of *wff* K.

When time is not critical, the system can use a *breadth-first, forward mode* processing to recompute the certainty measures of the modified *wffs*, attempting to restore the integrity of the rule deduction graph. In the case illustrated in the previous figure, this implies an update of fact L and rule R6 (both of which were not considered by the backward mode, since they did not play any role in determining the value of the proposed query, e.g. *wff* K).

The structure of the graph can also change, as new rule instances are created or deleted, due to changes in the facts' values, (as opposite to facts' certainty values). The deduction graph is updated and bad flags are propagated throughout the network

### 2.3.3 Rule Firing Control via Context Activation

A user-definable threshold can be attached to each rule context, either by local definition or by inheritance from a rule class. A rule context is defined as a conjunction of conditions that must be satisfied before the rule can be considered for premise evaluation. Each condition is described by a predicate on object-level *wffs* (facts in problem domain), or control-level *wffs* (markers asserted by meta-rules). The semantics of a context C attached to an inference rule (establishing the weak logical equivalence between A and B) is given by the following expression:

$$C \rightarrow ( A \leftrightarrow^{s,n} B )$$

where $s$ and $n$ indicate the lower bounds of the degree of sufficiency and necessity that the rule provides; $\rightarrow$ represents the strong material implication; $\leftrightarrow$ denotes the weak logical equivalence.

The context mechanism provides the following features:

1. By activating/deactivating subsets of the KB, it limits the number of rules that will be considered relevant at any given time, thus increasing the overall system efficiency.
2. By only considering the rules relevant to a given situation, it allows the knowledge engineer to effectively use the necessary conditions in the rule's premise. It is now possible to distinguish between the failure of a necessary test (described in the premise) and the failure of the rule's applicability (traditionally described by other clauses in the same premise and now explicitly represented in the context).
3. By using predicates on the control-level *wffs*, it provides the required programmability for defining flexible control strategies, such as causing sequences of rules to be executed, firing default rules, ordering and handling time-dependent information, etc.
4. By using hierarchical contexts, it can be used as an organizing principle for the knowledge acquisition task.

## 3. THE OBJECT BASED SIMULATION ENVIRONMENT

The second block of the MS/MT architecture is the simulation environment. This environment is centered around LOTTA, an object-oriented symbolic battle management simulator that maintains time-varying situations in a multi-player antagonistic game [4]. The development environment based on LOTTA constitutes a testbed for validating new techniques in reasoning with uncertainty and for performing information fusion functions [11]. The development environment is composed of four basic modules: the *window manager*, the *annotation system*, the *symbolic simulator (LOTTA)*, and the *Interface (KEELA)* . The simulation environment was used to program the naval scenario in which the information fusion and situation assessment tasks were performed.

### 3.1 The Information Fusion/Situation Assessment Problem

The Information Fusion (IF)/Situation Assessment (SA) requires a variety of tasks in which uncertainty pervades both the input data and the knowledge bases. Beside its intrinsic uncertainty, usually the information dealt in each task is also incomplete, time-varying, and, sometimes, erroneous. Thus, the SA problem represents a strong challenge for most automated reasoning systems, since it requires an integration of the uncertainty management with a truth maintenance system (belief revision system) to maintain the integrity of the inference base (or of its relevant subset). The SA problem also requires the reasoning system to detect useless and contradicting infonnation, rejecting the fonner and resolving the latter.

There is no uniformly agreed definition of what a situation assessment problem entails. The following definitions have been compiled and summarized from a variety of sources [5,8] to succinctly describe the SA problem. Given a platform (aircraft, ship, tank) in a potentially hostile environment, the process of performing Situation Assessment consists of the following tasks:

1. Sensor data must be collected from various sources and described as reports.
2. Time-stamped sensor reports must be consolidated into tracks (each track is the trace of an object followed by a given sensor).
3. Tracks associated to the same object must be fused into a platform.
4. The detected platform must be classified and identified (by class and type).
5. Node organization (formation of the identified platforms), use of special equipment, and maneuvering must be recognized.
6. Using the knowledge of the opponent's doctrines and rules of engagement, the recognized formation and observed use of special equipment must be explained by a probable intent, which is then translated into a threat assessment (retrospective SA).
7. This analysis is then projected into the future to evaluate plausible plans and to determine likely interesting developments of the current situation (prospective SA).

The first four tasks constitute what is generally known as Information Fusion and define the scope of the first MS/MT experiment.

### 3.2 Example of RUM rules

The RUM knowledge base (KB) used in MS/MT application is composed of approximately forty rules, each of which can be instantiated by new sensor reports, new tracks, or new platforms. A representative sample of such a KB is provided by the following two rules.

English Version of Rule-500 (identifying submarines):

*Assuming that a radar was used to generate a sensor report (that with other reports generated by the same sensor has been attached to a track associated with a platform), if the first time that the platform was detected (in the track's first report), the platform was located at a distance of at most twenty miles from our radar (i.e., it was a close-distance radar pop-up) then it is most likely that the platform is a submarine. Otherwise, there is a small chance that it is not a submarine.*

RUM's Version of the same rule:

```
(add-template 'sub.pos.id-close.pop.up-500                     ; Name
    'msmt                                                      ; KB
    '((u-lessp (get.uncertain.value (get.value ?track 'first.report) 'range)
       (fuzz 20)))                                             ; Premise-list
    '(((get.value ?track 'platform) class.name submarine s2.rules))  ; Consequence-list
    '((?track first.report))                                   ; List of wffs in premise
    '(?track)                                                  ; List of units in premise
    '((is-in-class? (get.value ?report 'track) 'source '(radar lotta)))  ; Context
    '(most.likely small.chance)                                ; Sufficiency and necessity
    't3                                                        ; Aggregation T-norm
    '(submarine track.templates))                              ; Rule class & instantiation templ.
```

English Version of Rule-550 (identifying submarines):

*Assuming that a sonar was used to generate a sensor report (that with other reports generated by the same sensor has been attached to a track associated with a platform), if the detected platform has a low noise emission, and is located at a depth of at least twenty meters, then it is extremely likely that it is a submarine. Otherwise, it may not be a submarine.*

RUM's Version of the same rule:

```
(add-template 'sub.pos.id-sonar-550                                  ; Name
    'msmt                                                            ; KB
    '((is-value? ?report 'noise-emissions 'low)                      ; Premise-list
      (u-lessp (get.uncertain.value ?report 'elevation) (fuzz -20)))
    '(((get.platform ?report) class.name submarine s2.rules))        ; Consequence-list
    '((?report elevation))                                           ; List of wffs in premise
    '(?report)                                                       ; List of units in premise
    '((is-in-class? (get.value ?report 'track) 'source '(sonar lotta))) ; Context
    '(extremely.likely it.may)                                       ; Sufficiency and necessity
    't3                                                              ; Aggregation T-norm
    '(submarine report.templates))                                   ; Rule class & instantiation templ.
```

### 3.2.1 Notes on the Calculi Selection for Rule 500 and 550

The T-norm used to detach the conclusion of rule 500 and 550 is $T_3$. This is due to the fact that we want to obtain the smallest certainty interval associated with the detached conclusion. The T-conorm used to aggregate the certainties of the detachments of both rules is $S_2$. This assignment indicates a lack of correlation among the two rules, which is substantiated by the fact that independent sources of information (radar and sonar) are used in the context of the two rules.

## 4. THE EXPERIMENT

In the experiment, a modified version of the naval situation assessment scenario used by NOSC to test STAMMER and STAMMER2 [1,7,9] was created. In this modified scenario, a missile cruiser of the type CGN36 operating with a surface radar (SPS 10) and a passive sensor (GPS-3) faced two platforms (selected from a set of possible platform classes such as cruisers, destroyers, frigates, patrol hydrofoils, submarines, merchant ships, and fishing boats). One of the two platforms was using an active sensor (navigational radar), while the second platform was not using any sensor.

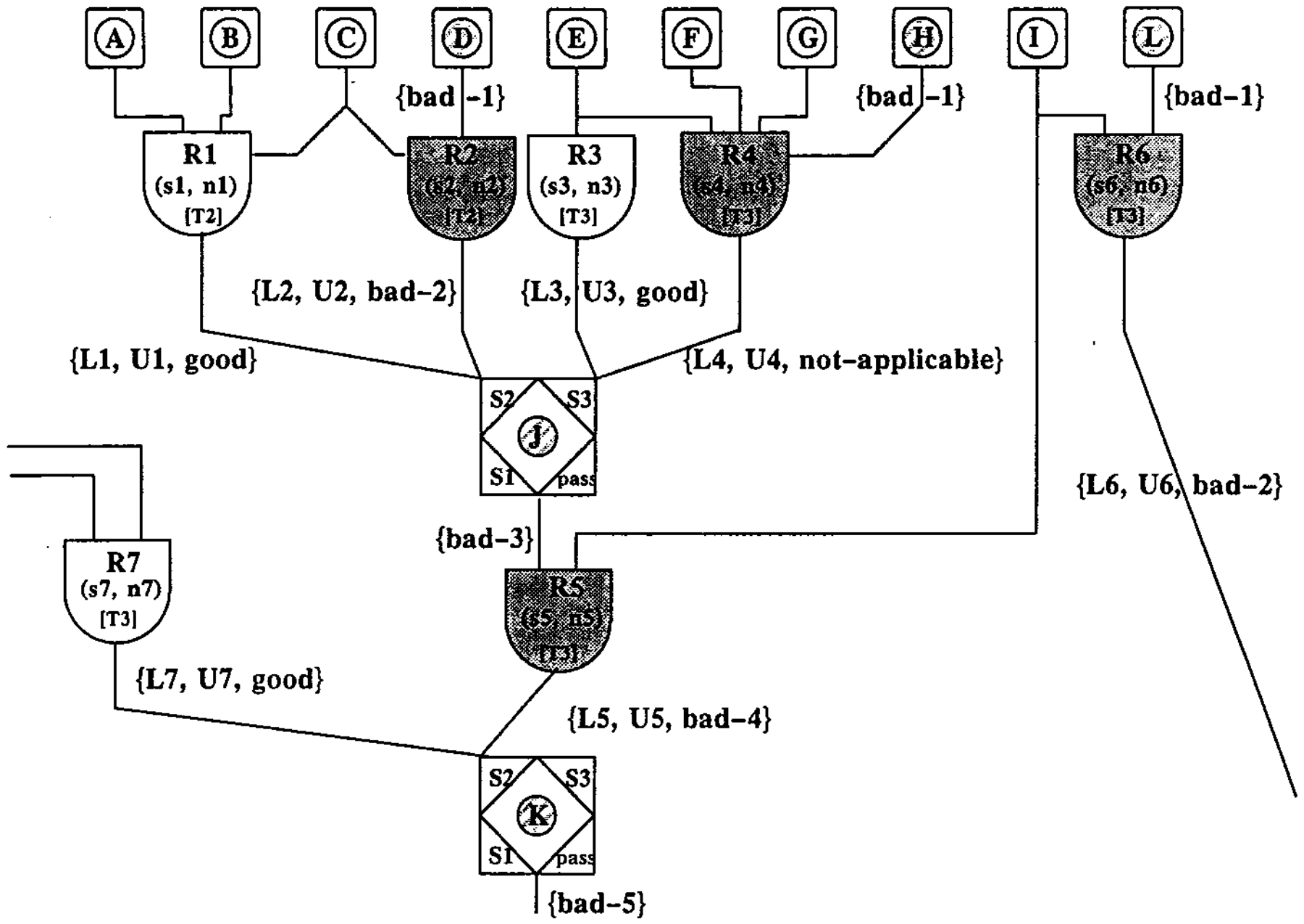

**Figure 4: Subgraph of the MSMT Knowledge Base**

The cruiser's task was to track, correlate, and classify each detected object. Both passive and active sensors on the cruiser were run twice, generating sensor reports which were translated through the KEELA interface into observed *wffs*. The sensor report information generated by the passive sensor (GPS-3) contained the heading, position, range, speed, and time at which the platform was detected. This information was attached to a track (TRACK-0) which maintained subsequent sensor reports generated by the same sensor and associated with the same platform (PLATFORM-439). Another track (TRACK-3) was generated by using a second sensor (SPS-10). The information from both tracks was attached to the same platform. Another track (TRACK-7), generated by the cruiser's active sensor, was attached to the second platform. Figure 4 illustrates a portion of the knowledge base where the report, track, and platform information is stored. In the same figure it is possible to observe the rule instantiation (by track) of the two rules (500 and 550) described in section 3.2.

The query posed to RUM was to deduce the class value of the first platform from the tracks information. Using the RUM knowledge base and the backward chaining mode, various attributes of the platform were inferred or observed. The platform was correctly identified as a merchant ship. This conclusion was based on the fact that the platform was reasonably close to a shipping lane, it was traveling at a a typical merchant's speed (in the 9-14 miles/hour range), it was not maneuvering, nor was it trying to dodge the cruise's surface radar. Figure 2 (used as an example in section 2.1.1 to describe an uncertainty unit attached to a *wff*) shows the uncertainty information and meta-information associated with the value assignments to the variable [*Platform-439 Class-name*]. In the slot VALUES, we can see the platform classes which were considered by the system and their corresponding certainty bounds: *Merchant* [.69 1], *Submarine* [0 .2], *Fishing Boat* [0 .02]

The best value in terms of its certainty is clearly the one which identifies Platform-439 as a *Merchant*. Its certainty's lower bound indicates a reasonably large amount of positive (confirming) evidence. Its upper bound indicates the absence of any negative (refuting) evidence. The class *Submarine* obtained no confirming evidence and a large amount of negative evidence. The refuting evidence was provided by rule 500, which from the failure to observe a *close-distance radar pop-up* determined that there was only a small chance for the platform to be a submarine. The class *Fishing Boat* also had no confirming evidence and an overwhelming amount of negative evidence. This refuting evidence was due to the fact that the platform was too far from the fishing areas, too big, and was using a radar (rules 340, 320, and 330). This information can be obtained from Figure 2, by observing the logical support for each of the three values considered for the *wff* [*Platform-439 Class-name*], and from Figure 5, by observing the dominant rules for each value. Each rule instance, fired to infer a value of the *wff*, has a cached certainty value (lower and upper bounds) and an associated validity flag. Thus, Figure 5 provides the information which was schematically described by the acyclic graph depicted in Figure 3.

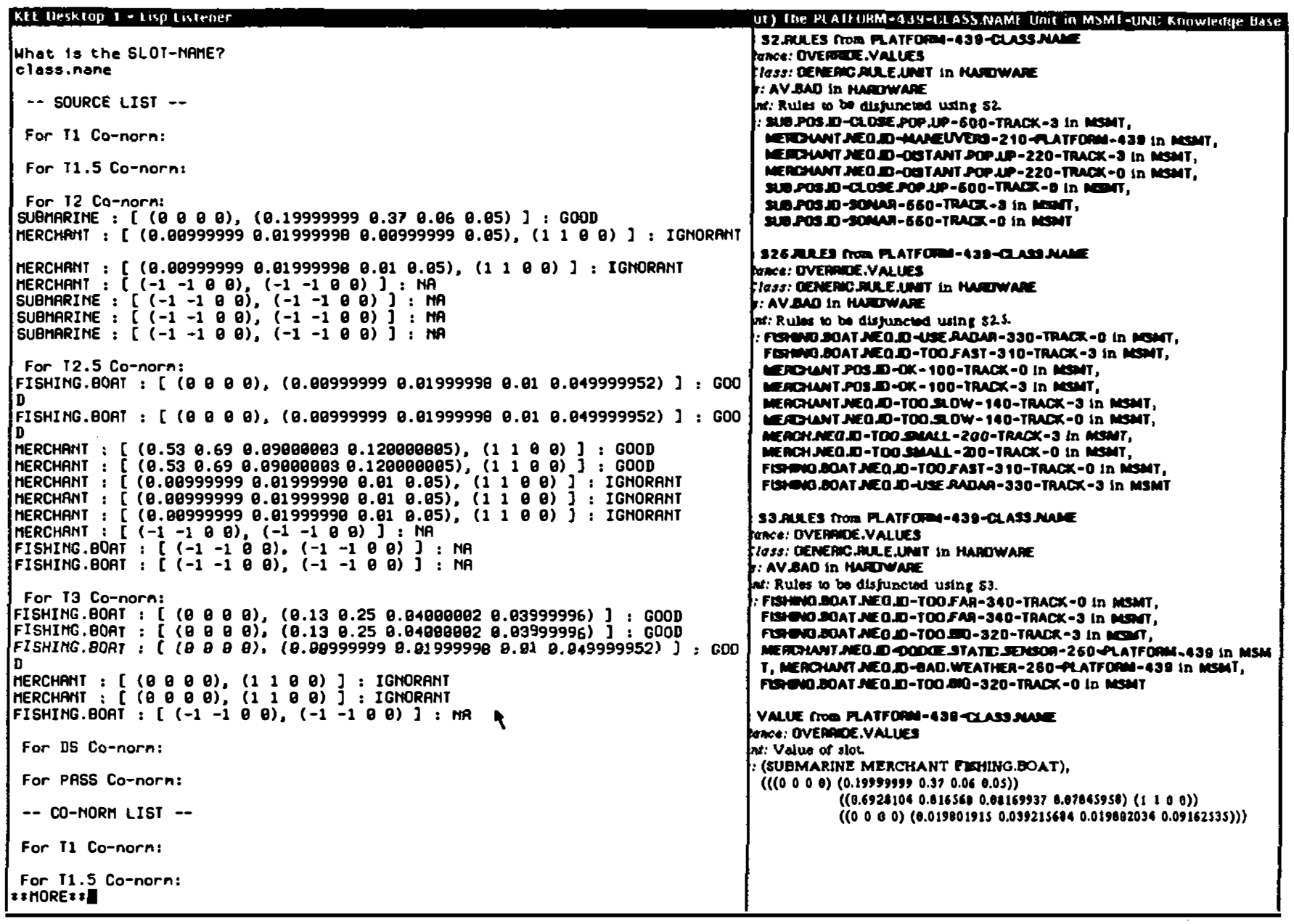

**Figure 5: Relevant Rule Instances in [*Platform-439 Class.name*] Logical Support**

## 5. REMARKS AND CONCLUSIONS

RUM's layered architecture properly addresses the requirements imposed by the SA problem. The representation layer captures the uncertain information about the *wffs* (lower and upper bounds) used by the calculi in the inference layer to determine the uncertainty of the conclusions. The representation layer also captures the uncertain meta-information (evidence source or logical support, measures of ignorance and conflict) used by the belief revision system and other mechanisms in the control layer.

The inference layer provides the knowledge engineer with a rich selection of well-understood calculi to properly represent existing correlations among rules. Numerical computations performed in this layer are efficiently implemented by using a four parameter representation for the uncertainty bounds, supported by a set of closed form formulae that implement the truth functional uncertainty calculi [2].

The control layer provides the explicit selection and modification of uncertainty calculi. Its context activation mechanism allows the reasoning system to focus on the relevant subsets of the changing inference base (the acyclic deductive graph). The uncertain-belief revision maintains the integrity of those relevant subsets, reflecting the changes of the information. RUM's development environment provides the traceability of *wffs* and rules that is required for proper KB development and refinment.

The MS/MT experiment described in this paper has been used to illustrate RUM's capabilities in an IF/SA application. It is a complete experiment, but certainly not a complex one. A more strenuous and realistic validation of RUM is in progress: currently RUM is successfully being used as the reasoning system of the Situation Assessment module in DARPA's Pilot's Associate Program [11]. In this application, the six tasks (described in section 4) that comprise the retrospective SA problem are addressed by RUM in Scenarios involving up to twenty platforms. This application is also used to derive some of the real-time requirements that will represent the focus of RUM's future development.

## 6. REFERENCES

[1] R.J. Bechtel, P.H. Morris, "STAMMER: System for Tactical Assessment of Multisource Messages, Even Radar", NOSC Technical Document 252, May 1979.

[2] P.P. Bonissone, & K.S. Decker, "Selecting Uncertainty Calculi and Granularity: An Experiment in Trading-off Precision and Complexity", in L.Kanal & J.Lemmer (Eds.) *Uncertainty in Artificial Intelligence*, pp. 217-247, North-Holland, 1986.

[3] P.P. Bonissone, S. Gans, K.S. Decker, "RUM: A Layered Approach to Reasoning with Uncertainty", to appear in the *Proceedings of the Tenth International Joint Conference on Artificial Intelligence (IJCAI-87)*, Milano, Italy, 1987.

[4] P.P. Bonissone, J.K. Aragones, K.S. Decker, "LOTTA: An Object Based Simulator for Reasoning in Antagonistic Situations", Working Paper, General Electric Corporate Research and Development, Schenectady, New York, 1987.

[5] P.P. Bonissone, "Summarizing and Propagating Uncertain Information with Triangular Norms", to appear in the *International Journal of Approximate Reasoning*, North-Holland, 1987.

[6] A. Clarkson, *Toward Effective Strategic Analysis: New Applications of Information Technology*, Westview Press, Boulder, Colorado, 1981.

[7] J. P. Ferranti, "Evaluation of the Artificial Intelligence Program STAMMER2 in the Tactical Situation Assessment Problem", M.S. Thesis, Naval Postgraduate School, Monterey, Ca., 1981.

[8] T.S. Levitt, G.J. Courand, M.R. Fehling, R.M. Fung, C.F. Kaun, R.M. Tong, "Intelligence Data Analysis", TR-1056-6, Volume 1, Advanced Decision Systems, Mountain View, CA., 1984.

[9] D.C. McCall, P.H. Morris, D.F. Kilber, R.J. Bechtel, "STAMMER2 Production System for Tactical Situation Assessment", NOSC Technical Document 298, October 1979.

[10] Schweizer, B., Sklar, A., "Associative Functions and Abstract Semi-Groups", *Publicationes Mathematicae Debrecen*, Vol. 10, pp. 69-81, 1963.

[11] L.M. Sweet, P.P. Bonissone, A.L. Brown, S. Gans, "Reasoning with Incomplete and Uncertain Information for Improved Situation Assessment in Pilot's Associate", Proceedings of the *12th DARPA Strategic Systems Symposium*, Monterey, California, October 28-30, 1986.